\documentclass[preprint,review,12pt]{elsarticle}

\usepackage{amssymb}

\usepackage{amsmath,amsfonts}
\usepackage{algorithmic}
\usepackage{algorithm}
\usepackage{array}
\usepackage[caption=false,font=normalsize,labelfont=sf,textfont=sf]{subfig}
\usepackage{textcomp}
\usepackage{stfloats}
\usepackage{url}
\usepackage{verbatim}
\usepackage{graphicx}

\usepackage{bm}
\usepackage{multirow}
\usepackage{arydshln}
\usepackage{overpic}
\usepackage{color}
\usepackage{makecell}

\def\ie{\emph{i.e.}}
\def\eg{\emph{e.g.}}

\def\etal{{\em et al.}}

\newcommand{\redbf}[1]{\textbf{\textcolor{red}{#1}}}
\newcommand{\bluebf}[1]{\textbf{\textcolor{blue}{#1}}}
\newcommand{\tableref}[1]{Table~\textbf{\textcolor{red}{\ref{#1}}}}
\newcommand{\figref}[1]{Fig.~\textbf{\textcolor{red}{\ref{#1}}}}

\usepackage{wrapfig} 
\usepackage{flushend} 

\journal{Pattern Recognition}

\begin{document}

\begin{frontmatter}

\title{Joint Super-Resolution and Inverse Tone-Mapping:
\\
A Feature Decomposition Aggregation Network and A New Benchmark}

\author[1,2]{Gang Xu}
\ead{gangxu@mail.nankai.edu.cn}

\author[1]{Yu-Chen Yang}
\ead{yycstat@mail.nankai.edu.cn}

\author[3]{Liang Wang}
\ead{wangliang@nlpr.ia.ac.cn}

\author[4]{Xian-Tong Zhen \corref{cor1}}
\ead{zhenxt@gmail.com}

\author[1]{Jun Xu}
\ead{nankaimathxujun@gmail.com}

\address[1]{School of Statistics and Data Science, Nankai University, Tianjin 300071, China}
\address[2]{College of Computer Science, Nankai University, Tianjin 300071, China}
\address[3]{National Lab of Pattern Recognition, Institute of Automation, CAS, Beijing, China}
\address[4]{Guangdong University of Petrochemical Technology, Maoming, Guangdong, China}

\cortext[cor1]{Corresponding author}

\begin{abstract}
Joint Super-Resolution and Inverse Tone-Mapping (joint SR-ITM) aims to increase the resolution and dynamic range of low-resolution and standard dynamic range images.
Recent networks mainly resort to image decomposition techniques with complex multi-branch architectures.
However, the fixed decomposition techniques would largely restricts their power on versatile images.
To exploit the potential power of decomposition mechanism, in this paper, we generalize it from the image domain to the broader feature domain.
To this end, we propose a lightweight Feature Decomposition Aggregation Network (FDAN).
In particular, we design a Feature Decomposition Block (FDB) to achieve learnable separation of detail and base feature maps, and develop a Hierarchical Feature Decomposition Group by cascading FDBs for powerful multi-level feature decomposition.
Moreover, for better evaluation, we collect a large-scale dataset for joint SR-ITM, \ie, SRITM-4K, which provides versatile scenarios for robust model training and evaluation.
Experimental results on two benchmark datasets demonstrate that our FDAN is efficient and outperforms state-of-the-art methods on joint SR-ITM.
The code of our FDAN and the SRITM-4K dataset are available at \url{https://github.com/CS-GangXu/FDAN}.
\end{abstract}




\end{frontmatter}


\section{Introduction}
The Ultra High Definition (UHD) and High Dynamic Range (HDR) display systems defined by Rec.2100~\cite{union2016recommendation} can well broadcast UHD-HDR images with a wider field of view (\eg, 4K or 8K resolution) and greater brightness/detail enjoyment (\eg, 10 or 12 pixel bit-depth), than the Full High Definition (FHD, 2K resolution) and Standard Dynamic Range (SDR, 8 pixel bit-depth) display systems defined by Rec.709~\cite{union2015recommendation}.
However, benefited from cheap photography and display devices like mobile phones, the abundant resources for FHD-SDR images somewhat limits the popularization of expensive UHD and HDR display systems~\cite{kim2019deep}.
To better display FHD-SDR images on UHD-HDR display systems, it is essential to convert the FHD-SDR images to genuine UHD-HDR format, which is the task of ``joint Super-Resolution and Inverse Tone-Mapping'' (joint SR-ITM)~\cite{kim2019deep,kim2020jsi}.

\begin{figure}[!t]
\centering
\includegraphics[width=13.8cm]{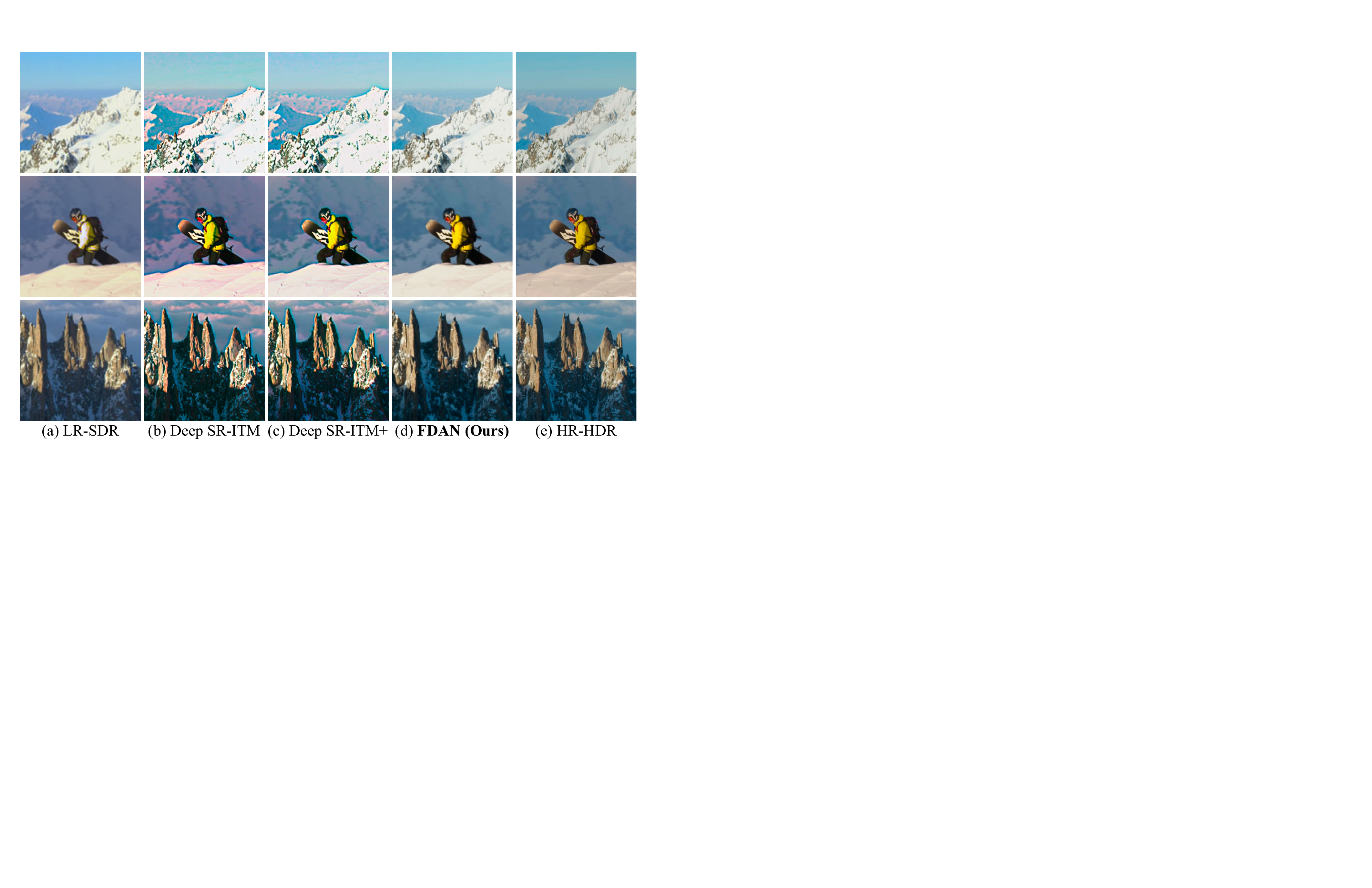}
\vspace{-9mm}
\caption{\textbf{Visual comparison of our FDAN and Deep SR-ITM~\cite{kim2019deep} on joint SR-ITM}. Deep SR-ITM is trained on the dataset provided in~\cite{kim2019deep}, while Deep SR-ITM+ and FDAN network are trained on our SRITM-4K dataset.}
\label{fig:intro}
\end{figure}

A direct solution for joint SR-ITM is to cascade existing super-resolution (SR) methods~\cite{dong2015image,kim2016accurate,zhang2018image} and inverse tone-mapping (ITM) operators~\cite{eilertsen2017hdr,liu2020single,WU2022108620}, or vice versa. These two-stage solutions convert low resolution (LR) FHD-SDR images to high-resolution (HR) UHD-HDR ones. However, these methods usually suffer from a huge computational burden coupled with extra amounts of parameters for feature extraction and reconstruction~\cite{kim2019deep,kim2020jsi}. To remedy this problem, several one-stage methods~\cite{kim2019deep,kim2020jsi} exploit the inherent correlation between high-frequency details and local contrasts in natural images, via decomposition techniques~\cite{he2012guided}. To this end, these methods~\cite{kim2019deep,kim2020jsi} first decompose the input LR-SDR image into the detail and base components via filtering techniques~\cite{he2012guided}, and then directly learn to enhance details and local contrasts simultaneously in an end-to-end manner. These decomposition-based methods achieve promising joint SR-ITM performance. However, the fixed filtering operation could hardly decompose versatile images very well and largely restricts the learning capability of these methods. Moreover, since the decomposition is performed directly on the input image, current one-stage methods~\cite{kim2019deep,kim2020jsi} have to tackle the decomposed detail and base components by multiple branches, with growing parameters and computational costs.


To better exploit the power of decomposition mechanism for joint SR-ITM task, in this paper, we design a novel Feature Decomposition Block (FDB) to learn data-driven feature decomposition, instead of performing fixed decomposition from the input images~\cite{kim2019deep,kim2020jsi}. The proposed feature-level decomposition has two key advantages over previous image-level counterparts~\cite{kim2019deep,kim2020jsi}: 1) it enjoys a larger receptive field by extending the decomposition from the input image to general feature maps; 2) it avoids the complex multi-branch network architectures in~\cite{kim2019deep,kim2020jsi}, and hence results in a single-branch network with high computational efficiency. To extract different scales of detail and contrast feature maps, we cascade several FDB blocks into a Hierarchical Feature Decomposition Group (HFDG). With our HFDG, we propose a lightweight Feature Decomposition Aggregation Network (FDAN) to progressively refine the learning of data-driven feature decomposition for joint SR-ITM. As shown in~\figref{fig:intro}(c) and~\figref{fig:intro}(d), both trained on our SRITM-4K dataset (will be introduced later), the proposed FDAN network outperforms previous work~\cite{kim2019deep} on the reconstructed HR-HDR images.

One possible limitation for the joint SR-ITM task is the small dataset collected in~\cite{kim2019deep}, which contains only 10 pairs of LR-SDR and HR-HDR videos. In this dataset, 39,840 pairs of LR-SDR ($40\times40$ or $80\times80$) and HR-HDR ($160\times160$) patches are extracted as the training set, and 28 pairs of LR-SDR ($960\times540$ or $1,920\times1,080$) and HR-HDR ($3,840\times2,160$) images are taken as the test set. However, the small training patches could hardly provide adequate receptive field of the trained networks for versatile test images, restricting their performance on joint SR-ITM. Besides, a set of 28 test images is relatively insufficient for conclusive evaluation.
To provide a large-scale dataset with adequate receptive field, in this paper, we collect a joint SR-ITM dataset, called SRITM-4K, consisting of 4K resolution images in diverse scenarios selected from 30 LR-SDR and HR-HDR video pairs.
%
%
It provides 5,000 pairs of HR-HDR ($3,840\times2,160$) images and LR-SDR images from $1,920\times1,080$ to $240\times135$ for $\times2\sim\times16$ SR tasks, respectively. For the test set, we also provide 200 pairs of HR-HDR and LR-SDR images (in the same setting as the training set) for comprehensive model evaluation. As shown in~\figref{fig:intro}(b) and~\figref{fig:intro}(c), the Deep SR-ITM~\cite{kim2019deep} trained on our SRITM-4K dataset achieves better results than that trained on the dataset in~\cite{kim2019deep}. Extensive experiments on our SRITM-4K dataset and the dataset in~\cite{kim2019deep} demonstrate that our FDAN network is efficient and outperforms state-of-the-art joint SR-ITM methods.
  
In summary, our main contributions are three-fold:
  \begin{itemize}
    \item We develop a novel Feature Decomposition Aggregation Network (FDAN) for joint SR-ITM, generalizing the fixed and specific image decomposition by guided filtering to flexible and general feature decomposition.
    \item To fulfill the gap between the small scale of the dataset in~\cite{kim2019deep} and the huge complexity of deep networks, we construct a new SRITM-4K dataset for joint SR-ITM, with $5,200$ high-resolution (4K) images in diverse scenarios.
    \item Experiments on these two datasets validate the advantage of our SRITM-4K dataset, demonstrating that our FDAN is efficient and outperforms previous methods.
  \end{itemize}

Our paper are organized as follows. In~\S\ref{sec:Related Work}, we introduce the related work.
In~\S\ref{sec:Proposed Approach}, we present the architecture of the proposed FDAN, including Feature Decomposition Block (FDB) and Hierarchical Feature Decomposition Group (HFDG).
Extensive experiments on our SRITM-4K dataset and the dataset in~\cite{kim2019deep} are conducted in~\S\ref{sec:Experiments}, which demonstrate the advantages of our FDAN over state-of-the-art joint SR-ITM methods. We conclude our paper in~\S\ref{sec:Conclusion}.

\section{Related Work}
\label{sec:Related Work}
 \noindent
  \textbf{Image Super-Resolution} (SR) aims to recover the high-resolution (HR) images from the corresponding low-resolution (LR) ones.
  Early SR methods mainly resort to hand-crafted regression models~\cite{zeyde2010single,yang2010image,timofte2014a+}.
  SRCNN~\cite{dong2015image} is among the first SR convolutional neural networks (CNNs).
  Later, residual and dense connections~\cite{he2016deep,huang2017densely,LI2021107610} are incorporated into several representative SR networks~\cite{kim2016accurate,FU2019780,ARUN2019431,AHN2022108649}.
  To achieve more visually pleasing performance, Ledig~\etal~\cite{ledig2017photo} tackled SR under the adversarial learning framework~\cite{goodfellow2014generative}.
  Recently, attention mechanisms are also widely utilized by SR networks. For example, channel attention~\cite{hu2018squeeze} is utilized in~\cite{zhang2018image,BEHJATI2023108997}, while self-attention~\cite{vaswani2017attention,wang2018non} is exploited by~\cite{mei2021image,chen2021pre}.
  For real-world applications, several lightweight SR methods~\cite{ahn2018fast,zhao2020efficient,luo2020latticenet} are developed to reduce the amount of parameters or computational costs.
  To this end, Ahn~\etal~\cite{ahn2018fast} introduced a cascaded residual network architecture to reduce the number of parameters.
  Feature distillation is utilized in~\cite{hui2019lightweight,liu2020residual} to extract discriminative features. Zhao~\etal~\cite{zhao2020efficient} introduced a pixel attention network of only $\sim$270K parameters, but with huge computational costs.
  Unfortunately, these SR networks could not directly enhance the image dynamic range.
  In this work, we propose the FDAN network with a lightweight structure to perform joint SR-ITM.
  
  \noindent
  \textbf{Inverse tone-mapping} (ITM) aims to reconstruct high dynamic range (HDR) images from the low or standard dynamic range (LDR or SDR) ones.
  Early ITM methods~\cite{banterle2006inverse,banterle2007framework,rempel2007ldr2hdr} mainly estimate an expand map to guide the dynamic range expansion.
  Several CNNs are utilized for data-driven HDR reconstruction~\cite{eilertsen2017hdr,wu2018deep,WU2022108620}.
  For example, end-to-end autoencoder frameworks are incorporated by Eilertsen~\etal~\cite{eilertsen2017hdr} and Wu~\etal~\cite{wu2018deep} to hallucinate plausible details of the HDR image.
  Spatial alignment is also exploited in~\cite{kalantari2017deep,chen2021hdr,KHAN2023109344,YAN2022108342} to merge multiple LDR images with different exposures for HDR reconstruction.
  Santos~\etal~\cite{santos2020single} proposed a feature masking mechanism to adaptively adjust the weight of features in the saturated region.
  The reversion of the image formation pipeline is also studied in~\cite{liu2020single,chen2021new} to produce HDR images.
  Here, Liu~\etal~\cite{liu2020single} reversed the LDR formation pipeline with three specialized CNNs.
  Chen~\etal~\cite{chen2021new} proposed a three-step solution pipeline according to the formation pipeline for SDR-to-HDR content translation.
  However, these methods are unable to enhance the spatial resolution of the input images.
  In this paper, our FDAN network is feasible to increase both spatial resolution and dynamic ranges for UHD display.
  
  \noindent
  \textbf{Joint SR-ITM} aims to enhance both the spatial resolution and the dynamic range of an input image.
  %
  %
  Kim~\etal~\cite{kim2019deep} decomposed the input image into detail and base components via filtering techniques~\cite{he2012guided}, and learned to simultaneously enhance the detail and local contrast.
  To obtain images of high perceptual quality, Kim~\etal~\cite{kim2020jsi} trained the network under the generative framework~\cite{goodfellow2014generative}.
  However, the performance of these networks upon versatile scenarios is largely limited by the fixed decomposition~\cite{he2012guided} performed only on the input image.
  Also, tackling the decomposed base and detail components individually by corresponding branches comes with great computational costs.
  To alleviate these problems, we generalize the decomposition scheme from image domain to general feature domain, and propose a Feature Decomposition Block (FDB) for learnable feature separation.
  To exploit multi-scale information, we cascade several FDB blocks into a group, with which we propose a Feature Decomposition Aggregation Network for efficient joint SR-ITM.

\begin{figure*}[!t]
\centering
\includegraphics[width=13.8cm]{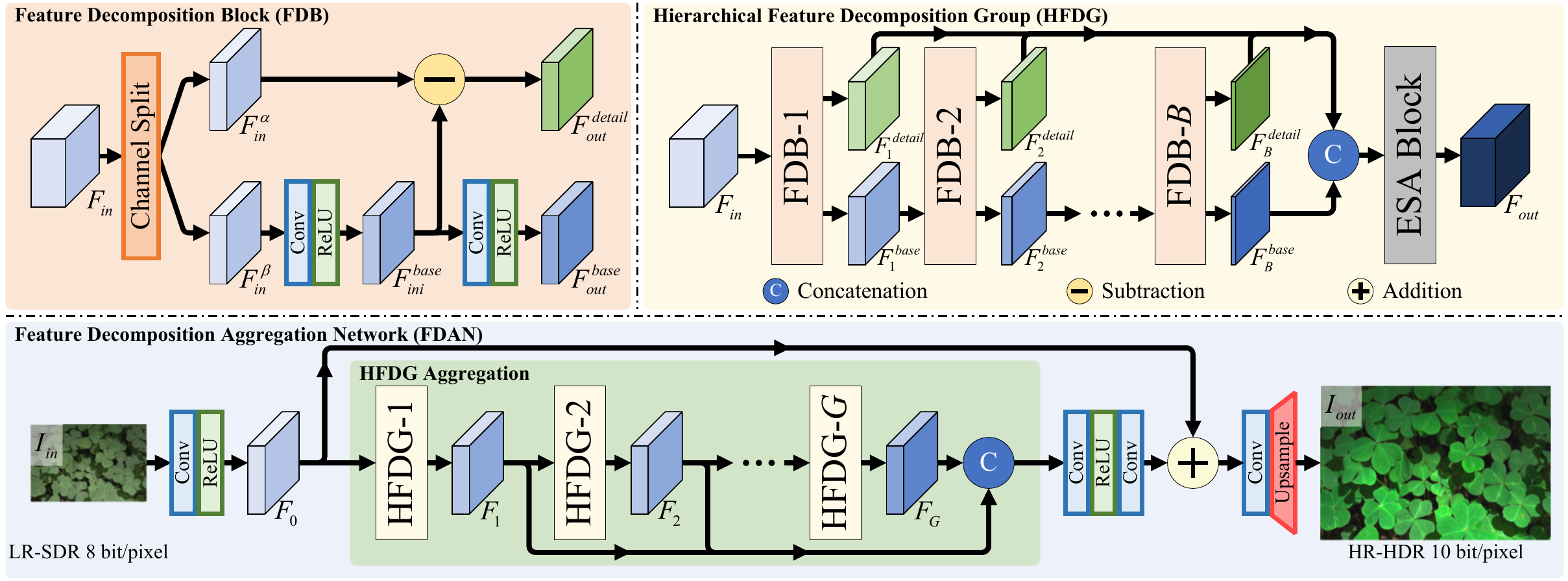}
\vspace{-9mm}
\caption{
      \textbf{Illustration of our Feature Decomposition Aggregation Network (FDAN) for joint SR-ITM}.
      \textbf{Top-left}: our FDB splits the input feature $\bm{F}_{in}$ into $\bm{F}_{in}^{\alpha}$ and $\smash{\bm{F}_{in}^{\beta}}$ along the channel dimension and extract the initial base feature $\bm{F}_{ini}^{base}$. Then we subtract the initial base feature $\bm{F}_{ini}^{base}$ from $\bm{F}_{in}^{\alpha}$ to generate the output detail feature $\bm{F}_{out}^{detail}$, and refine the initial base feature $\bm{F}_{ini}^{base}$ to output the enhanced base feature $\bm{F}_{out}^{base}$.
      \textbf{Top-right}: our HFDG hierarchically combines $B$ FDBs to extract the multi-level detail and base feature maps, and adopt an Enhanced Spatial Attention (ESA) block to refine the concatenation of all the detail features and the $B$-th base feature.
      \textbf{Bottom}: given the LR-SDR image $\bm{I}_{in}$, we extract the initial feature and feed it into the HFDG Aggregation consisting of $G$ cascaded HFDG groups to extract features with different scales of receptive fields. Then we aggregate all the features from $G$ HFDG groups and add the initial feature to the resulting feature. Finally, the HR-HDR image $\bm{I}_{out}$ is obtained by a convolutional layer and a Pixel Shuffle layer.
}
\label{fig:method_pipeline}
\end{figure*}

\section{Proposed Method}
\label{sec:Proposed Approach}
In this section, we first introduce our Feature Decomposition Block (FDB) in~\S\ref{sec:block} and the Hierarchical Feature Decomposition Group (HFDG) in~\S\ref{sec:group}, which learn fine-grained feature separation and multi-level feature fusion, respectively.
Then, we present our Feature Decomposition Aggregation Network (FDAN) for joint SR-ITM in~\S\ref{sec:network}.
At the end, the implementation details are provided in~\S\ref{sec:implementation}.

\subsection{Feature Decomposition Block}
\label{sec:block}

Previous one-stage joint SR-ITM methods~\cite{kim2019deep,kim2020jsi} first decompose an input LR-SDR image $\bm{I}_{in}\in\mathbb{R}^{H\times W \times 3}$, where $H$ and $W$ are the height and width, into the base and detail components via guided image filtering~\cite{he2012guided}, and enhance them separately by different network branches.
However, the decomposition~\cite{he2012guided} is performed only on the input image with fixed hyperparameters, which is not robust to versatile scenarios.
To generalize the decomposition from the specific image domain to the general feature domain, we propose a novel Feature Decomposition Block (FDB) to perform the separation of feature details and contrasts in a data-driven manner. The pipeline of the feature decomposition block is as shown in~\figref{fig:method_pipeline} (top-left).


The goal of our FDB block is to extract detail and base feature maps from the input feature $\bm{F}_{in}\in\mathbb{R}^{H\times W\times C}$ (\eg, extracted from the input LR-SDR image $\bm{I}_{in}$) for joint SR-ITM, where $C$ is the number of channels.
A straightforward way to this goal is to individually perform base feature extraction from the whole feature map $\bm{F}_{in}$, and obtain the detail feature by subtracting the base feature from $\bm{F}_{in}$.
%
%
To improve the computational efficiency, we propose to split the input feature $\bm{F}_{in}$, along the channel dimension, into two feature maps $\bm{F}_{in}^{\alpha}$ and $\smash{\bm{F}_{in}^{\beta}}$, both in $\mathbb{R}^{H\times W\times C/2}$.
This also makes our FDB block lightweight for efficient model design.
%
Then we extract the initial base feature $\bm{F}_{ini}^{base}\in\mathbb{R}^{H\times W\times C/2}$ from the feature map $\smash{\bm{F}_{in}^{\beta}}$ by a $1\times1$ convolutional layer, followed by a $3\times3$ convolutional layer to output the enhanced base feature $\bm{F}_{out}^{base}\in\mathbb{R}^{H\times W\times C/2}$ with larger receptive field.
To extract and output the detail feature $\bm{F}_{out}^{detail}$, we subtract the initial base feature $\bm{F}_{ini}^{base}$ from the feature map $\bm{F}_{in}^{\alpha}$ by
\begin{equation}
\setlength\abovedisplayskip{1pt}
\setlength\belowdisplayskip{1pt}
  \begin{aligned}
      \bm{F}_{out}^{detail} = \bm{F}_{in}^{\alpha} - \bm{F}_{ini}^{base}.
  \end{aligned}
\end{equation}
In this way, our FDB block decomposes the input feature map $\bm{F}_{in}$ into the detail feature map and base feature map in a data-driven manner:
\begin{equation}
\setlength\abovedisplayskip{1pt}
\setlength\belowdisplayskip{1pt}
  \begin{aligned}
  \text{FDB}(\bm{F}_{in})=[\bm{F}_{out}^{base}, \bm{F}_{out}^{detail}].	\end{aligned}
\end{equation}
Next, we stack multiple FDB blocks into a cascade to exploit the power of multi-level decomposition for fine-grained feature extraction and separation.

\subsection{Hierarchical Feature Decomposition Group}
\label{sec:group}

To obtain fine-grained detail feature map and base feature map, we cascade several (\eg, $B$) FDB blocks into a Hierarchical Feature Decomposition Group (HFDG) to separate the input feature $\bm{F}_{in}$ into different levels of base and detail feature maps, as shown in~\figref{fig:method_pipeline} (top-right).
Given the input feature $\bm{F}_{in}\in\mathbb{R}^{H\times W\times C}$, we employ the first FDB block (called FDB-1) to decompose it into the base feature $\bm{F}_{1}^{base}$ and detail feature $\bm{F}_{1}^{detail}$, both in $\mathbb{R}^{H\times W\times C/2}$.
Then the base feature $\bm{F}_{1}^{base}$ is fed into the subsequent FDB block for next-level feature decomposition.
In general, the $b$-th FDB block ($b\in\{2,...,B\}$) decomposes the base feature $\bm{F}_{b-1}^{base}\in\mathbb{R}^{H\times W\times C/2^{b-1}}$ from previous FDB block as
\begin{equation}
\setlength\abovedisplayskip{1pt}
\setlength\belowdisplayskip{1pt}
\begin{aligned}
\text{FDB}_{b}(\bm{F}_{b-1}^{base})=[\bm{F}_{b}^{base}, \bm{F}_{b}^{detail}],
\end{aligned}
\end{equation}
where $\bm{F}_{b}^{base}$ and $\bm{F}_{b}^{detail}$ (both in $\mathbb{R}^{H\times W\times C/2^{b}}$) are the output base and detail feature maps, respectively.

\begin{figure}[!t]
\centering
\includegraphics[width=8.8cm]{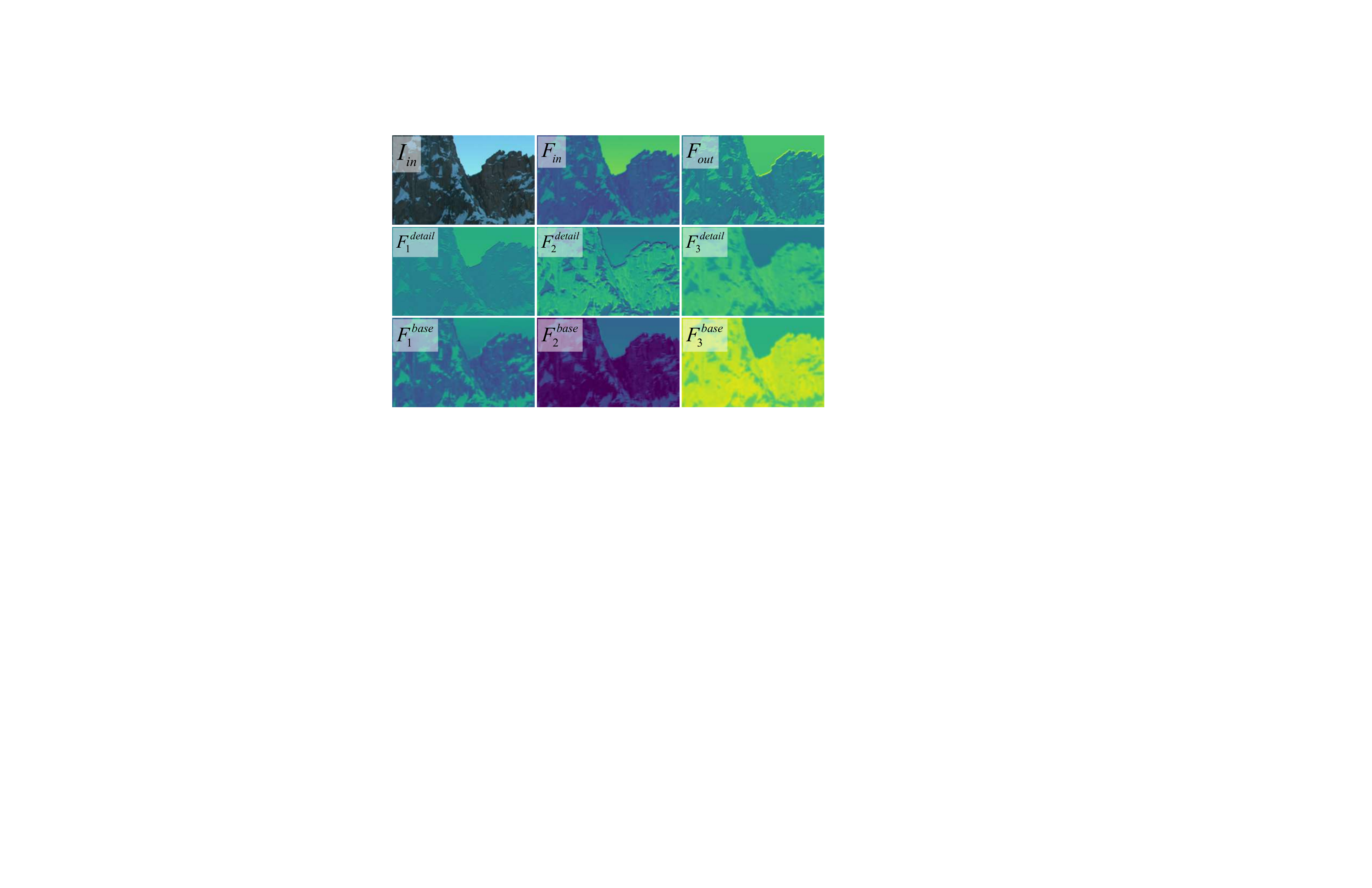}
\caption{
\textbf{Visualization of intermediate features in HFDG}.
\textbf{1-st row}: the input image, the 9-th and 18-th channel feature maps of the input and output features.
\textbf{2-nd row}: the 4-th, 1-st, and 1-st channel feature maps of the output detail features obtained from 1-st to 3-rd FDB, respectively.
\textbf{3-rd row}: the 5-th, 10-th, and 5-th channel feature maps of the output base features obtained from 1-st to 3-rd FDB, respectively.
}
\label{fig:method_visualization}
\end{figure}


Thus far, we have multiple hierarchically decomposed base and detail feature maps.
To well exploit their power for joint SR-ITM, we concatenate these features and refine them by a spatial attention mechanism.
Since the base feature $\bm{F}_{b-1}^{base}$ ($b=2,...,B$) have been decomposed into next-level detail and base features, we only concatenate all the detail features $\bm{F}_{b}^{detail}$ ($b=1,2,...,B$) and the $B$-th base feature $\bm{F}_{B}^{base}$ to avoid redundancy on information and computational costs.
Instead of the plain concatenation, here we choose to rescale the hierarchical feature maps from the FDBs and adaptively highlight the important regions in each hierarchical feature map according to the spatial distribution inside.
Thus, we employ an Enhanced Spatial Attention (ESA) block introduced in~\cite{liu2020rfan} for spatial refinement aggregation:
\begin{equation}
\setlength\abovedisplayskip{1pt}
\setlength\belowdisplayskip{1pt}
  \begin{aligned}
        \bm{F}_{out} = \text{ESA}([\bm{F}_{1}^{detail}, \bm{F}_{2}^{detail},..., \bm{F}_{B}^{detail} ,\bm{F}_{B}^{base}]),
  \end{aligned}
\end{equation}
where $[\cdot]$ denotes the concatenation operation and $\bm{F}_{out}$ is the output by HFDG group.
Note that $\bm{F}_{out}\in\mathbb{R}^{H\times W\times C}$ is of the same size as the input feature $\bm{F}_{in}$.
Considering that the ESA block~\cite{liu2020rfan} is not our contribution and not a key design in our main network, we put the detailed description of the ESA block~\cite{liu2020rfan} in the \textsl{supplementary material}.

In~\figref{fig:method_visualization}, we visualize the intermediate feature maps extracted by one HFDG with $B=3$ FDBs on the input LR-SDR image $\bm{I}_{in}$.
From $\bm{I}_{in}$, we extract the input feature $\bm{F}_{in}$, and decompose it into different levels of detail features $\bm{F}_{b}^{detail}$ and base features $\bm{F}_{b}^{base}$ ($b=1,2,3$).
As we can see, the three base features become coarser and coarser, and contain more and more structural information.
This is reasonable since each base feature $\bm{F}_{b}^{base}$ is fused by a $3\times3$ convolutional layer on that in the previous level.
Similarly, by subtracting the base feature from that in the previous level, three detail features also become coarser and coarser.
%
Finally, the multi-level detail and base features are integrated to produce an enhanced feature $\bm{F}_{out}$ with richer details and contrasts.
This shows that our HFDG effectively integrates different levels of detail and structural base features.
%

\subsection{Overall Network}
\label{sec:network}

With our HFDG group, we now develop a Feature Decomposition Aggregation Network (FDAN) for joint SR-ITM.
The overall architecture is shown in~\figref{fig:method_pipeline} (bottom).
Given the input LR-SDR image $\bm{I}_{in}\in\mathbb{R}^{H\times W\times 3}$, our FDAN network first extracts the initial feature $\bm{F}_{0}\in\mathbb{R}^{H\times W\times C}$ by a $3\times3$ convolutional layer.\ Then, we cascade $G$ HFDG groups to extract the features $\bm{F}_{g}\in\mathbb{R}^{H\times W\times C}$ ($g\in\{1,...,G\}$) with different scales of receptive fields.
The feature extracted by the $g$-th HFDG group (denoted as $\text{HFDG}_{g}$) is:
\begin{equation}
\setlength\abovedisplayskip{1pt}
\setlength\belowdisplayskip{1pt}
  \begin{aligned}
  \bm{F}_{g} 
  &= \text{HFDG}_{g}(\text{HFDG}_{g-1}(,...,\text{HFDG}_{1}(\bm{F}_{0}))).	\end{aligned}
\end{equation}
To fully exploit the multi-scale information for effective joint SR-ITM, we aggregate all these $G$ features $\{\bm{F}_{1},...,\bm{F}_{G}\}$ by direct concatenation, and integrate them by one $1\times1$ convolutional layer and one $3\times3$ convolutional operation.
The resulting feature is added by the initial feature $\bm{F}_{0}$ via a long skip connection, to utilize the low-level detail information for visual-pleasing reconstruction.
This can also stabilize and accelerate the training process of our FDAN network~\cite{kim2016accurate}.
Finally, we reconstruct the HR-HDR image $\bm{I}_{out}\in\mathbb{R}^{sH\times sW\times3}$ by a $3\times3$ convolutional operation and a Pixel Shuffle operation~\cite{shi2016real} with the scale factor $s$.

\begin{figure*}[t]
\centering
\includegraphics[width=13.8cm]{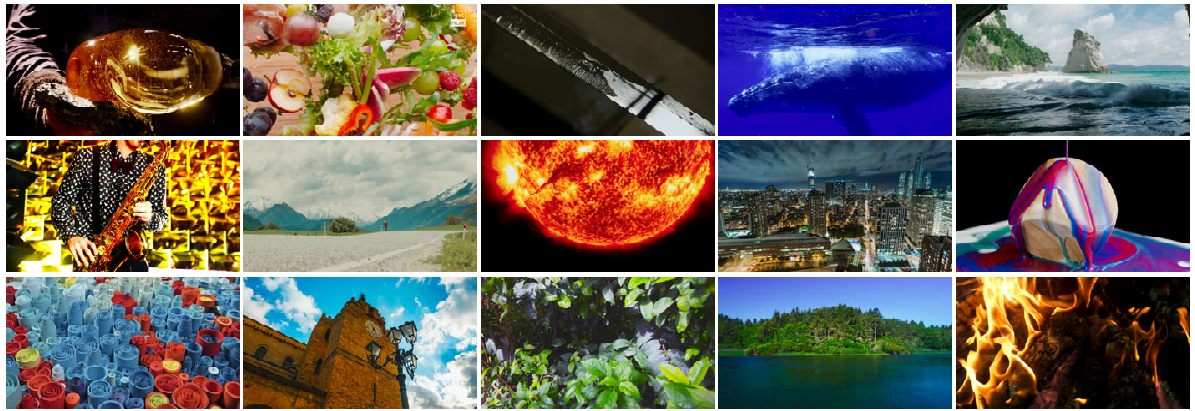}
\caption{
\textbf{Typical scenarios of the 30 HR-HDR videos in our SRITM-4K dataset.
}}
\label{fig:dataset_display}
\end{figure*}

\subsection{Implementation Details}
\label{sec:implementation}

Unless otherwise specified, in all experiments, we set the hyperparameters $C$, $B$, and $G$ in our FDAN as $C=48$, $B=3$, and $G=6$.
Each convolutional layer contains a convolutional operation and a ReLU nonlinear function.
We initialize our FDAN by the Kaiming initialization~\cite{he2015delving} without a pre-trained model.
We utilize the Adam optimizer~\cite{kingma2014adam} with $\beta_{1} = 0.9$ and $\beta_{2} = 0.999$ to optimize our FDAN with the $\ell_{1}$ loss function. On the dataset introduced in the next section, we train our FDAN for 1,200 epochs (a total of 93,760 iterations) with the batch size of 64 for the scale factor of 4, 8 and 16 and the batch size of 32 for the scale factor of 2 due to the GPU memory limitation.
The learning rate is initialized as $5\times10^{-5}$ and decayed to $1\times10^{-11}$ by a cosine annealing scheme~\cite{loshchilov2016sgdr} for every 120 epochs (9376 iterations).
The training takes about 6.75 hours on a RTX 2080Ti GPU with 11GB memory.

\section{Our SRITM-4K Dataset}
\label{sec:dataset}

\noindent
\textbf{Motivation}.
The work of~\cite{kim2019deep} provides a joint SR-ITM dataset with 39,840 training patches of size $160\times160$ extracted from 7 videos, and 28 test images.\ The joint SR-ITM networks trained on this dataset suffer from small receptive field, and thus could not well process high-resolution images in real-world scenarios.\ To advance the research on joint SR-ITM, we collect a new large-scale dataset with 5,200 images in 4K resolution, called ``SRITM-4K''.

\begin{figure}[t]
  \centering
  \begin{overpic}[width=8.8cm]{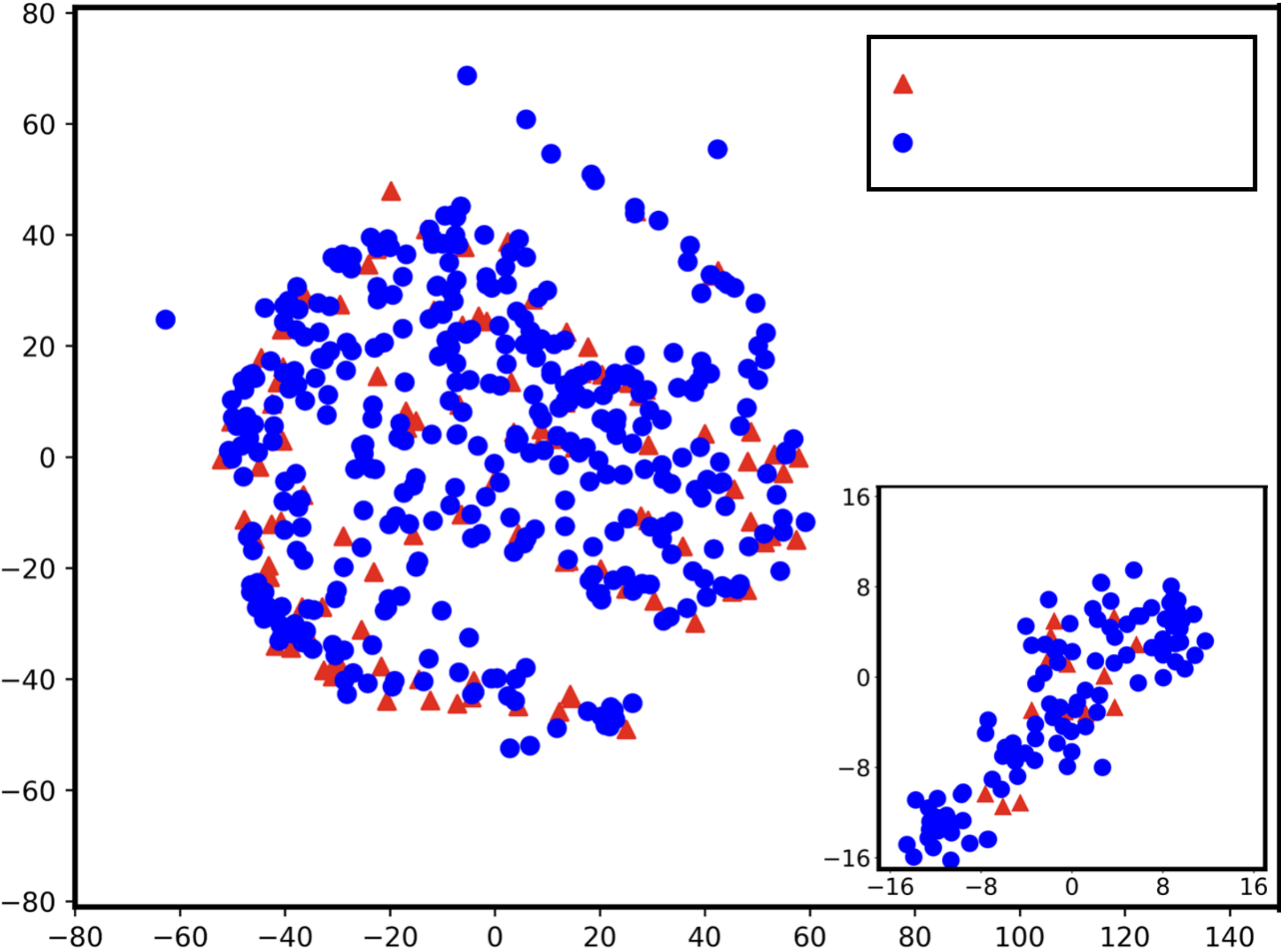}
  \put(72.0, 66.8){\tiny{Dataset in~\cite{kim2019deep}}}
  \put(72.0, 62.2){\tiny{SRITM-4K}}
  \put(7.5, 70.0){\scriptsize{Training sets}}
  \put(69.3, 32.9){\scriptsize{Test sets}}
  \end{overpic}
  \vspace{-3mm}
  \caption{
  \textbf{Sample distribution} of our SRITM-4K dataset (\textcolor{blue}{blue} points) and the dataset in~\cite{kim2019deep} (\textcolor{red}{red} triangles) using t-SNE~\cite{van2008visualizing}.
  Our SRITM-4K dataset provides more diversity than the dataset proposed by Kim~\etal~\cite{kim2019deep}, which could help to fully exploit the potential of network for joint SR-ITM.
  }
  %
  \label{fig:dataset_visualization}
\end{figure}
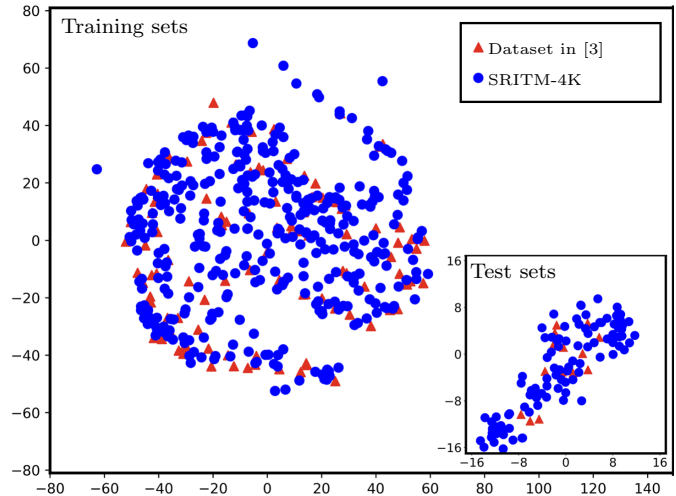

\noindent
\textbf{Dataset construction}.\ We collect 30 pairs of HR-SDR (Rec.709 display format~\cite{union2015recommendation}) and HR-HDR (Rec.2100 display format~\cite{union2016recommendation}) videos in 4K resolution ($3840\times2160$) from Youtube.\ The bit depths of HR-SDR and HR-HDR videos are 8 bit/pixel and 10 bit/pixel, respectively.
We uniformly sample 5,000 pairs of HR-SDR and HR-HDR images from 25 pairs of corresponding videos as the \texttt{training} set, and 200 pairs of HR-SDR and HR-HDR images from the rest 5 pairs of corresponding videos as the \texttt{test} set.
To show the diversity of our SRITM-4K dataset, we visualize the typical scenarios of 30 HR-HDR videos in our SRITM-4K dataset, in~\figref{fig:dataset_display}.
%
To generate the LR-SDR images, we downsample the HR-SDR images by bicubic interpolation with scale factors of 2, 4, 8, and 16.
In this way, we obtain the LR-SDR images of sizes $1,920\times1,080$, $960\times540$, $480\times270$, or $240\times135$, for $\times$2, $\times$4, $\times$8, or $\times$16 joint SR-ITM, respectively.
Similar to~\cite{kim2019deep}, the images in our SRITM-4K dataset are encoded in the YUV color space.\\
%
%
\noindent
\textbf{Discussion}.
Compared with the previous dataset in~\cite{kim2019deep}, our SRITM-4K dataset not only contains a larger amount of training and test images in diverse scenarios, but also provides original images (rather than cropped patches) to train networks with larger receptive field.
To quantitatively compare with the dataset in~\cite{kim2019deep}, we employ the t-SNE~\cite{van2008visualizing} to visualize the sample distribution of our SRITM-4K dataset and the dataset in~\cite{kim2019deep} in an embedded 2D space.
For the training set, if we directly visualize all the samples from the two datasets, the samples would be overwhelmed by each other.
To avoid this problem, we instead select proper amounts, but in equal ratio, of samples from the two datasets for better visualization.
Since the contents of different frames are roughly similar in one video, we set the ratio of the sample amounts selected from two datasets as $25:7$, which is the ratio between the amounts of source videos in our dataset and that in~\cite{kim2019deep}.
Specifically, we randomly select 375 samples and 105 samples from the training sets of our SRITM-4K dataset and the dataset in~\cite{kim2019deep}.
For the test set, we take all the test images into consideration.
As shown in~\figref{fig:dataset_visualization}, the training (or test) set in our SRITM-4K dataset is distributed in a more diverse yet comprehensive manner than that in~\cite{kim2019deep}.
This validates the advantage of our SRITM-4K dataset over~\cite{kim2019deep} for the joint SR-ITM task.
More details can be found in the \textsl{Supplementary File}.
%
%

\section{Experiments}
\label{sec:Experiments}

\subsection{Experimental Setup}

\noindent
\textbf{Dataset}.
We conduct experiments on our SRITM-4K dataset and the dataset in~\cite{kim2019deep}.
For SRITM-4K dataset, we use the \texttt{training} set and \texttt{test} set in our SRITM-4K dataset as the training and evaluation datasets, respectively.
In every iteration during the training process, we randomly crop one patch of size $256\times256$ from each HR-HDR image as the ground truth.
We also crop one patch, with the same location as that in the HR-HDR image, from the corresponding HR-SDR image, and downsample it by bicubic interpolation with a scale factor of $2$, $4$, $8$ or $16$.
The obtained LR-SDR patches are of sizes $128\times128$, $64\times64$, $32\times32$ or $16\times16$, respectively.
Note that although here we crop the training patches of $256\times256$, it is feasible to crop bigger patches for consideration of larger receptive fields.
The cropped patches are horizontally flipped and randomly rotated with $90^{\circ}$, $180^{\circ}$, or $270^{\circ}$ for data augmentation.
For dataset in~\cite{kim2019deep}, we use the \texttt{training} set and \texttt{test} set proposed in~\cite{kim2019deep} as the training and evaluation datasets, respectively.
During the training process, we adopt the officially released HR-HDR patches of size $160\times160$ as the ground truth.
The corresponding LR-SDR patches of size $40\times40$ or $80\times80$ with the scale factor of $2$ or $4$ are regarded as the input.
Following previous works~\cite{kim2019deep,kim2020jsi}, we do not apply data augmentation when training methods on dataset in~\cite{kim2019deep}.

\noindent
\textbf{Evaluation metrics}.
To evaluate the performances of networks for joint SR-ITM, we employ five metrics: PSNR, SSIM~\cite{wang2004image}, HDR-VDP3~\cite{mantiuk2011hdr}, PU-PSNR~\cite{azimi2021pu21} and PU-SSIM~\cite{azimi2021pu21}.
Specifically, PSNR and SSIM~\cite{wang2004image} are calculated on the Y-Channel of the YUV color space. 
The number of PU-PSNR~\cite{azimi2021pu21} is computed on the linearized RGB channels, while HDR-VDP3~\cite{mantiuk2011hdr} and PU-SSIM~\cite{azimi2021pu21} are calculated on the linearized luminance channel.
More details on the evaluation settings are provided in the \textsl{Supplementary File}.

%
%

\subsection{Comparison on Our SRITM-4K Dataset}

\noindent
\textbf{Comparison methods}.
We compare our FDAN network with state-of-the-art super-resolution (SR) methods~\cite{zhang2018image,liu2020residual}, cascaded two-stage SR-ITM methods~\cite{liu2020residual,chen2021hdrunet}, and one-stage SR-ITM methods~\cite{kim2019deep,kim2020jsi} for joint SR-ITM.
For SR methods, we compare with the RCAN~\cite{zhang2018image} and lightweight RFDN~\cite{liu2020residual}.
For two-stage SR-ITM methods, we adopt the RFDN~\cite{liu2020residual} for SR and the HDRUNet~\cite{chen2021hdrunet} for ITM, under the ``first SR then ITM'' (SR+ITM) and ``first ITM then SR'' (ITM+SR) frameworks.
For one-stage SR-ITM methods, we compare with the Deep SR-ITM~\cite{kim2019deep} and JSI-GAN~\cite{kim2020jsi}.
For the comparison methods, we use their official codes for our experiments, except that for Deep SR-ITM~\cite{kim2019deep} we use the PyTorch implementation reproduced in~\cite{greatwallet2019}.
All these methods are retrained on our SRITM-4K \texttt{training} set and evaluated on our SRITM-4K \texttt{test} set.\\
\noindent
\textbf{Objective results}.
The quantitative comparison results among 4 scale factors are listed in~\tableref{table:exp_sota}.
One can see that, with a lightweight structure, our FDAN network achieves efficient performance on joint SR-ITM.
Specifically, compared with all other methods, our FDAN network enjoys the least space complexity (\eg amount of parameters and activations~\cite{zhang2020aim}) and the lowest computational costs (\eg number of FLOPs and MACs).
Besides, our FDAN outperforms the other methods, by at least 0.03dB, 0.35dB, 0.24dB, and 0.26dB in terms of PSNR at scale factors of 2, 4, 8, and 16, respectively.
On SSIM~\cite{wang2004image}, our FDAN achieves comparable performances with other methods.
%
On speed, our FDAN network runs at 195.34, 185.47, 198.63, and 197.92 frames-per-second (FPS) at the scale factors of 2, 4, 8, and 16.
All these results demonstrate that our FDAN network is very efficient and effective for joint SR-ITM task at 4K resolution.\\

\begin{table*}[!h]
  \begin{center}
  \vspace{-0.2cm}
  \setlength\tabcolsep{0.5pt}
  \caption{
  \textbf{Comparison of different methods on the number of parameters (Params), Activations~\cite{zhang2020aim}, FLOPs, MACs, PSNR, SSIM~\cite{wang2004image}, HDR-VDP3~\cite{mantiuk2011hdr}, PU-PSNR~\cite{azimi2021pu21}, and PU-SSIM~\cite{azimi2021pu21} by our \textbf{SRITM-4K} dataset}.
  ``$\uparrow$'' or ``$\downarrow$'' means that larger or smaller is better.
  The best, second best and third best results are highlighted in \redbf{red}, \bluebf{blue} and \textbf{bold}, respectively.
  To avoid the metrics from saturating, we only utilize the luminance information in Y-Channel to calculate the evaluation metrics of PSNR and SSIM.
  The PU-PSNR~\cite{azimi2021pu21} is calculated on the linearized RGB channels, while HDR-VDP3~\cite{mantiuk2011hdr} and PU-SSIM~\cite{azimi2021pu21} are computed on the linearized luminance channel.
  }
  \vspace{-0.5cm}
  \resizebox{\textwidth}{!}{
  \begin{tabular}{@{\hskip 2pt}c@{\hskip 2pt}|c|@{\hskip 0pt}r@{\hskip 10pt}|@{\hskip 0pt}r@{\hskip 24pt}|@{\hskip 0pt}r@{\hskip 10pt}|@{\hskip 0pt}r@{\hskip 10pt}|c|c|c|c|c}
  \hline
  Scale & Method & \multicolumn{1}{c|}{Params (K)$\downarrow$}  & \multicolumn{1}{c|}{Activations (G)$\downarrow$}   & \multicolumn{1}{c|}{FLOPs (G)$\downarrow$} & \multicolumn{1}{c|}{MACs (G)$\downarrow$} & PSNR (dB)$\uparrow$ & SSIM$\uparrow$ & HDR-VDP$\uparrow$ & PU-PSNR (dB)$\uparrow$ & PU-SSIM$\uparrow$ \\
  \hline
  \multirow{7}{*}{\rotatebox[origin=c]{0}{$\times 2$}}
  &RCAN~\cite{zhang2018image} & 8695.81 & 41.43 & 35762.56 &17883.38 & 30.46 & 0.9772 & 8.80 & 28.95 & \redbf{0.9719} \\
  &RFDN~\cite{liu2020residual} & \bluebf{566.36}  & \bluebf{4.85}    & \bluebf{2223.74} & \bluebf{1111.57} & \textbf{31.45} & \textbf{0.9779} & \bluebf{8.94} & \bluebf{29.99} & \bluebf{0.9663} \\
  &RFDN~\cite{liu2020residual} + HDRUNet~\cite{chen2021hdrunet} & 2217.85 & 21.17 & 8098.03 & 4047.20 & 31.32 & 0.9775 & \redbf{8.94} & 29.41 & 0.9582 \\
  &HDRUNet~\cite{chen2021hdrunet} + RFDN~\cite{liu2020residual} & 2217.85 & 8.94 & \textbf{3713.38} & \textbf{1852.66} & \bluebf{32.02} & \bluebf{0.9814} & \textbf{8.92} & \textbf{29.64} & \textbf{0.9641} \\
  \cdashline{2-11}[0.8pt/2pt]
  &Deep SR-ITM~\cite{kim2019deep} & 1863.42 & 7.05 & 7741.86 & 3879.45  & 29.19 & 0.9418 & 7.01 & 25.78 & 0.9127 \\
  &JSI-GAN~\cite{kim2020jsi} & \textbf{1454.15} & \textbf{5.54} & 6138.78 & 3066.72 & 25.75 & 0.9212 & 7.34 & 20.81 & 0.8622 \\
  &\textbf{FDAN (Ours)} & \redbf{126.66} & \redbf{2.31} & \redbf{404.44} & \redbf{200.62}  & \redbf{32.05}  & \redbf{0.9815} & 8.90 & \redbf{30.14} & 0.9628 \\
  \hline
  \multirow{7}{*}{\rotatebox[origin=c]{0}{$\times 4$}}
  &RCAN~\cite{zhang2018image} & 8778.95 & 10.77 & 9303.94 & 4658.08 & 29.89  & 0.9614 & 8.79 & 28.98 & 0.9352 \\
  &RFDN~\cite{liu2020residual} & \bluebf{581.95} & \bluebf{1.23} & \bluebf{572.03} & \bluebf{285.52} & 31.36  & \textbf{0.9702} & \textbf{8.82} & 29.13 & \bluebf{0.9523} \\
  &RFDN~\cite{liu2020residual} + HDRUNet~\cite{chen2021hdrunet} & 2233.44 & 17.55 & 6446.32 & 3221.15 & \textbf{31.39} & \bluebf{0.9714} & \bluebf{8.87} & \bluebf{29.19} & \textbf{0.9520} \\
  &HDRUNet~\cite{chen2021hdrunet} + RFDN~\cite{liu2020residual} & 2233.44 & \textbf{2.25} & \textbf{950.97} & \textbf{470.78} & \bluebf{31.40} & \redbf{0.9723} & \redbf{8.90} & \textbf{29.17} & \redbf{0.9530} \\
  \cdashline{2-11}[0.8pt/2pt]
  &Deep SR-ITM~\cite{kim2019deep} & \textbf{2011.13} & 2.31 & 2573.67 & 1282.45  & 29.35 & 0.9502 & 7.03 & 26.93 & 0.9189  \\
  &JSI-GAN~\cite{kim2020jsi} & 3025.90 & 2.85 & 3190.03 & 1593.95 & 31.15 & 0.9532 & 8.49 & 28.61 & 0.9259  \\
  &\textbf{FDAN (Ours)} & \redbf{142.24} & \redbf{0.59} & \redbf{117.22} & \redbf{58.21} & \redbf{31.75}  & 0.9693 & 8.77 & \redbf{29.82} & 0.9488 \\
  \hline
  \multirow{7}{*}{\rotatebox[origin=c]{0}{$\times 8$}}
  &RCAN~\cite{zhang2018image} & 15740.07 & 4.13 & 4762.08 & 2383.52 & 29.92  & 0.9463 & \textbf{8.55} & \bluebf{28.31} & 0.9180  \\
  &RFDN~\cite{liu2020residual} & \bluebf{644.30} & \bluebf{0.32} & \bluebf{159.12} & \bluebf{79.44} & 30.09 & 0.9477 & 8.46 & 28.08 & \textbf{0.9266} \\
  &RFDN~\cite{liu2020residual} + HDRUNet~\cite{chen2021hdrunet} & 2295.79 & 16.65 & 6033.41 & 3015.07 & \bluebf{30.40} & \redbf{0.9509} & \redbf{8.61} & \textbf{28.29} & \redbf{0.9280} \\
  &HDRUNet~\cite{chen2021hdrunet} + RFDN~\cite{liu2020residual} & 2295.79 & \textbf{0.58} & \textbf{254.77} & \textbf{125.81} & \textbf{30.36} & \bluebf{0.9508} & \bluebf{8.58}  & 28.23 & \bluebf{0.9273} \\
  \cdashline{2-11}[0.8pt/2pt]
  &Deep SR-ITM~\cite{kim2019deep} & \textbf{2158.85} & 1.12 & 1283.90 & 637.37 & 27.62  & 0.9261 & 7.15 & 25.44 & 0.9027 \\
  &JSI-GAN~\cite{kim2020jsi} & 9312.89 & 2.18  & 2452.85 & 1225.37 & 27.32 & 0.8156 & 6.30 & 24.13 & 0.7109 \\
  &\textbf{FDAN (Ours)} & \redbf{204.60} & \redbf{0.16} & \redbf{45.42} & \redbf{22.61} & \redbf{30.64}  & \textbf{0.9480} & 8.52 & \redbf{28.79} & 0.9231  \\
  \hline
  \multirow{7}{*}{\rotatebox[origin=c]{0}{$\times 16$}}
  &RCAN~\cite{zhang2018image} & 15887.79 & 1.58 & 1823.58 & 911.79  & 28.29 & 0.9301 & 8.01 & 26.56 & 0.9019  \\
  &RFDN~\cite{liu2020residual} & \bluebf{893.71} & \bluebf{0.10} & \bluebf{55.90} & \bluebf{27.92}  & 28.86 & 0.9293 & 8.01 & \textbf{27.07} & \redbf{0.9052}  \\
  &RFDN~\cite{liu2020residual} + HDRUNet~\cite{chen2021hdrunet} & 2545.20 & 16.42 & 5930.19 & 2963.55  & \textbf{28.96}  & \bluebf{0.9334} & \redbf{8.05}  & 27.04 & 0.9046  \\
  &HDRUNet~\cite{chen2021hdrunet} + RFDN~\cite{liu2020residual} & 2545.20 & \textbf{0.16} & \textbf{80.08} & \textbf{39.54}  & \bluebf{29.08}  & \redbf{0.9343} & \textbf{8.02} & \bluebf{27.12} & \textbf{0.9047}  \\
  \cdashline{2-11}[0.8pt/2pt]
  &Deep SR-ITM~\cite{kim2019deep} & \textbf{2306.56} & 0.83 & 952.52 & 475.86  & 26.09  & 0.9107 & 7.07 & 23.78 & 0.8826  \\
  &JSI-GAN~\cite{kim2020jsi} & 34460.86 & 2.01 & 2268.60 & 1133.86 & 24.54 & 0.5324 & 6.25 & 19.75 & 0.6325  \\
  &\textbf{FDAN (Ours)} & \redbf{454.00} & \redbf{0.06} & \redbf{27.48} & \redbf{13.71} & \redbf{29.34} & \textbf{0.9328} & \bluebf{8.03} & \redbf{27.62} & \bluebf{0.9050}  \\
  \hline
  \end{tabular}
  }
  \label{table:exp_sota}
  \end{center}
  \vspace{-0.8cm}
\end{table*}

\begin{figure*}[!t]
  \centering
    \includegraphics[width=13.8cm]{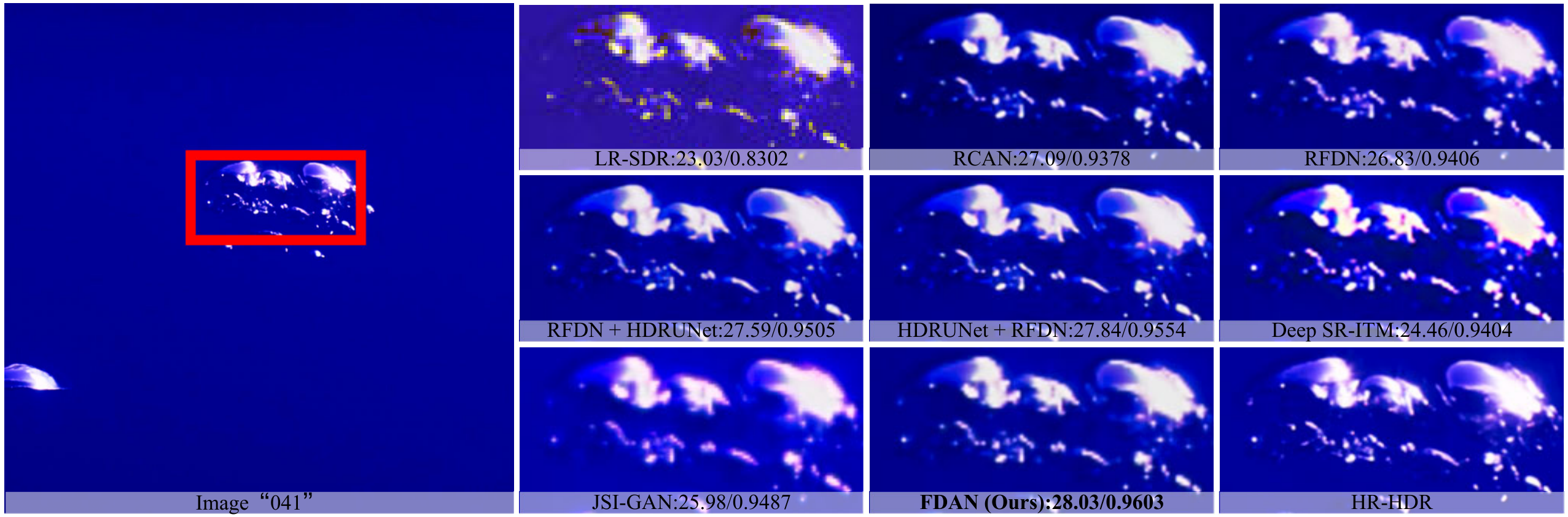}\\
    \includegraphics[width=13.8cm]{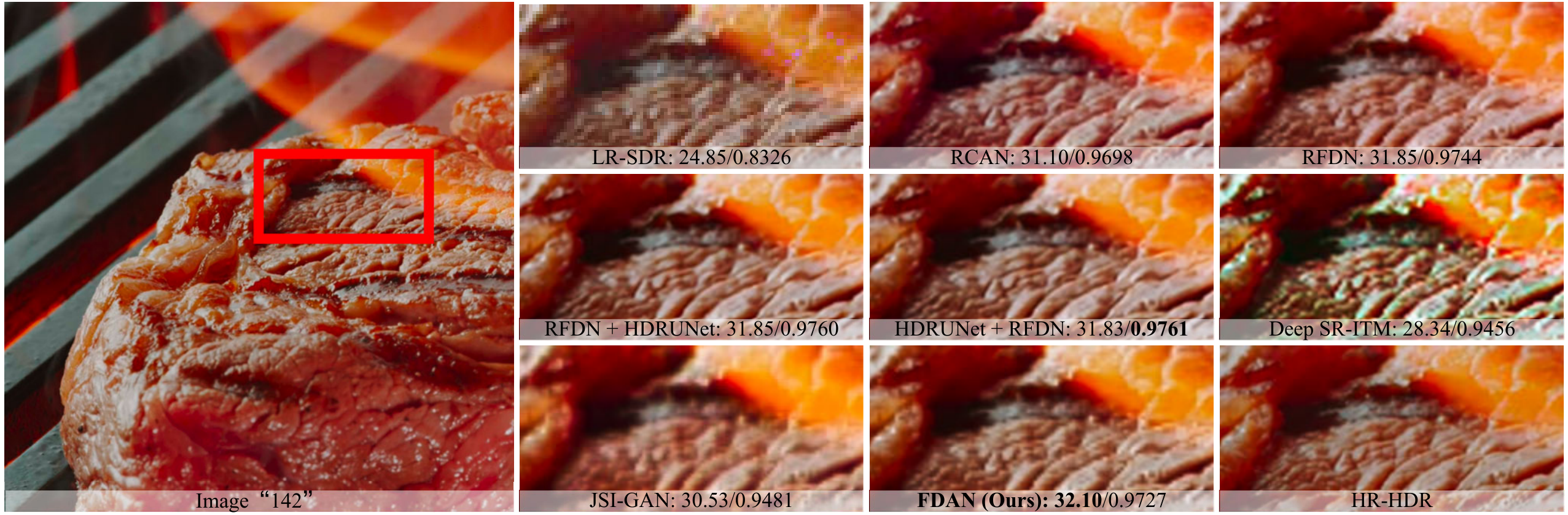}
      \caption{
      \textbf{Comparison of visual quality and PSNR (dB)/SSIM results by different joint SR-ITM methods on SRITM-4K with the scale factor of 4}.
      }
    \label{fig:experiment_visualization}
\end{figure*}

\noindent
\textbf{Visual quality}.
Here, we visualize the joint SR-ITM results by different methods with the scale factor of 4, following the pipeline provided in~\cite{kim2019deep}.
From~\figref{fig:experiment_visualization}, one can see that, owing to our feature decomposition and hierarchical feature extraction mechanism, our FDAN produces better visual results than those of the competitors.
%
For example, in the 1-st row, we observe that our FDAN outputs a clear background, while other SR-ITM methods bring artifacts and/or shadow in background.
In the 2-nd and 3-rd rows, our FDAN also achieves better reconstruction quality than the other methods, in terms of detail and structure.
More comparison results are provided in the \textsl{Supplementary File}.

\subsection{Comparisons on the Dataset in~\cite{kim2019deep}}
\label{sec:comparison_on_dsi}
\noindent
\textbf{Comparison methods}.
We also compare our FDAN network with the one-stage SR-ITM methods: Deep SR-ITM~\cite{kim2019deep} and JSI-GAN~\cite{kim2020jsi}, on the dataset in~\cite{kim2019deep}.
All these comparison methods are trained on the training set and evaluated on the test set of the dataset in~\cite{kim2019deep} according to the official settings, with the scale factors of 2, 4.
We adopt the PyTorch implementation reproduced in~\cite{greatwallet2019} for Deep SR-ITM~\cite{kim2019deep}, while using the official codes for JSI-GAN~\cite{kim2020jsi}.\\

\noindent
\textbf{Objective results}.
The quantitative comparison results are listed in~\tableref{table:exp_sota_dsi}.
One can see that our FDAN outperforms the other methods, by at least 0.11db and 0.26dB in terms of PSNR at scale factors of 2 and 4, respectively.
In terms of SSIM, our FDAN outperforms the other methods, by at least 0.0017 at a scale factor of 4 and is slightly inferior to some other methods at a scale factor of 2.
Note that our FDAN also enjoys the least space complexity (\eg, parameters and activations~\cite{zhang2020aim}) and the lowest computational costs (\eg, FLOPs and MACs) among the comparison methods.
%

Besides the official settings of LR-SDR and HR-HDR images with the scale factors of 2 and 4 provided in~\cite{kim2019deep}, we conduct additional SR-ITM settings with the scale factors of 8 and 16, to comprehensively evaluate the comparison methods on dataset in~\cite{kim2019deep}.
To obtain the LR-SDR images with the scale factors of 8 and 16, we downsample the LR-SDR images at the scale factor of 2 by a further scale factor of 4 and 8, respectively, through the standard bicubic interpolation.
All models are retrained to achieve their corresponding best performance.
The results of different competitors are listed in~\tableref{table:exp_sota_dsi}.
One can see that our FDAN still outperforms the other one-stage SR-ITM methods~\cite{kim2019deep,kim2020jsi} in terms of PSNR and SSIM, but with the least parameters (model complexity) and lowest FLOPs (computational costs).
This demonstrates the efficiency and effectiveness of our FDAN network over previous methods~\cite{kim2019deep,kim2020jsi} on joint SR-ITM.

\begin{table*}[!h]
  \vspace{-0.2cm}
  \begin{center}
  \setlength\tabcolsep{11pt}
  \caption{
  \textbf{Comparison of different methods on the number of parameters (Params), Activations~\cite{zhang2020aim}, FLOPs, MACs, PSNR, and SSIM~\cite{wang2004image} by the dataset in~\cite{kim2019deep}}.
  ``$\uparrow$'' (or ``$\downarrow$'') means that larger (or smaller) is better.
  The best, second best and third best results are highlighted in \redbf{red}, \bluebf{blue} and \textbf{bold}, respectively.
  To avoid the metrics from saturating, we only utilize the luminance information in Y-Channel to calculate the evaluation metrics of PSNR and SSIM.
  }
  \vspace{-0.5cm}
  \resizebox{\textwidth}{!}{
  \begin{tabular}{@{\hskip 2pt}c@{\hskip 2pt}|c|@{\hskip 0pt}r@{\hskip 20pt}|@{\hskip 0pt}c@{\hskip 0pt}|@{\hskip 0pt}r@{\hskip 20pt}|@{\hskip 0pt}r@{\hskip 18pt}|c|c}
  \hline
  Scale & Method & \multicolumn{1}{c|}{Params (K)$\downarrow$}  & \multicolumn{1}{c|}{Activations (M)$\downarrow$}   & \multicolumn{1}{c|}{FLOPs (G)$\downarrow$} & \multicolumn{1}{c|}{MACs (G)$\downarrow$} & PSNR (dB)$\uparrow$ & SSIM$\uparrow$ \\
  \hline
  \multirow{3}{*}{\rotatebox[origin=c]{0}{$\times 2$}}
  &Deep SR-ITM~\cite{kim2019deep} & \textbf{1863.42} & \textbf{7.05} & \textbf{7741.86} & \textbf{3879.45}  & \textbf{32.63} & \textbf{0.9200}   \\
  &JSI-GAN~\cite{kim2020jsi} & \bluebf{1454.15} & \bluebf{5.54} & \bluebf{6138.78} & \bluebf{3066.72} & \bluebf{32.83} & \redbf{0.9255} \\
  &\textbf{FDAN (Ours)} & \redbf{126.66} & \redbf{2.31} & \redbf{404.44} & \redbf{200.62}  & \redbf{32.94}  & \bluebf{0.9236}  \\
  \hline
  \multirow{3}{*}{\rotatebox[origin=c]{0}{$\times 4$}}
  &Deep SR-ITM~\cite{kim2019deep} & \bluebf{2011.13} & \bluebf{2.31} & \bluebf{2573.67} & \bluebf{1282.45}  & \bluebf{30.37} & \textbf{0.8447}    \\
  &JSI-GAN~\cite{kim2020jsi} & \textbf{3025.90} & \textbf{2.85} & \textbf{3190.03} & \textbf{1593.95} & \textbf{30.11} & \bluebf{0.8454} \\
  &\textbf{FDAN (Ours)} & \redbf{142.24} & \redbf{0.59} & \redbf{117.22} & \redbf{58.21} & \redbf{30.63}  & \redbf{0.8471}  \\
  \hline
    \multirow{3}{*}{\rotatebox[origin=c]{0}{$\times 8$}}
    & Deep SR-ITM~\cite{kim2019deep} & \bluebf{2158.85} & \bluebf{1.12} & \bluebf{1283.90} & \bluebf{637.37} & \textbf{27.25} & \textbf{0.7648} \\
    & JSI-GAN~\cite{kim2020jsi} & \textbf{9312.89} & \textbf{2.18}  & \textbf{2452.85} & \textbf{1225.37} & 27.12 & 0.7632 \\
    & \textbf{FDAN (Ours)} & \redbf{204.60} & \redbf{0.16} & \redbf{45.42} & \redbf{22.61} & \bluebf{27.40} & \bluebf{0.7722} \\
    \hline
    \multirow{3}{*}{\rotatebox[origin=c]{0}{$\times 16$}}
    & Deep SR-ITM~\cite{kim2019deep} & \bluebf{2306.56} & \bluebf{0.83} & \bluebf{952.52} & \bluebf{475.86} & 25.19 & 0.7314 \\
    & JSI-GAN~\cite{kim2020jsi} & \textbf{34460.86} & \textbf{2.01} & \textbf{2268.60} & \textbf{1133.86} & \textbf{25.32} & \textbf{0.7406} \\
    & \textbf{FDAN (Ours)} & \redbf{454.00} & \redbf{0.06} & \redbf{27.48} & \redbf{13.71} & \bluebf{25.56} & \bluebf{0.7434} \\
    \hline
  \end{tabular}
  }
  \label{table:exp_sota_dsi}
  \end{center}
  \vspace{-0.8cm}
\end{table*}

\begin{figure*}[t]
\centering
\includegraphics[width=13.8cm]{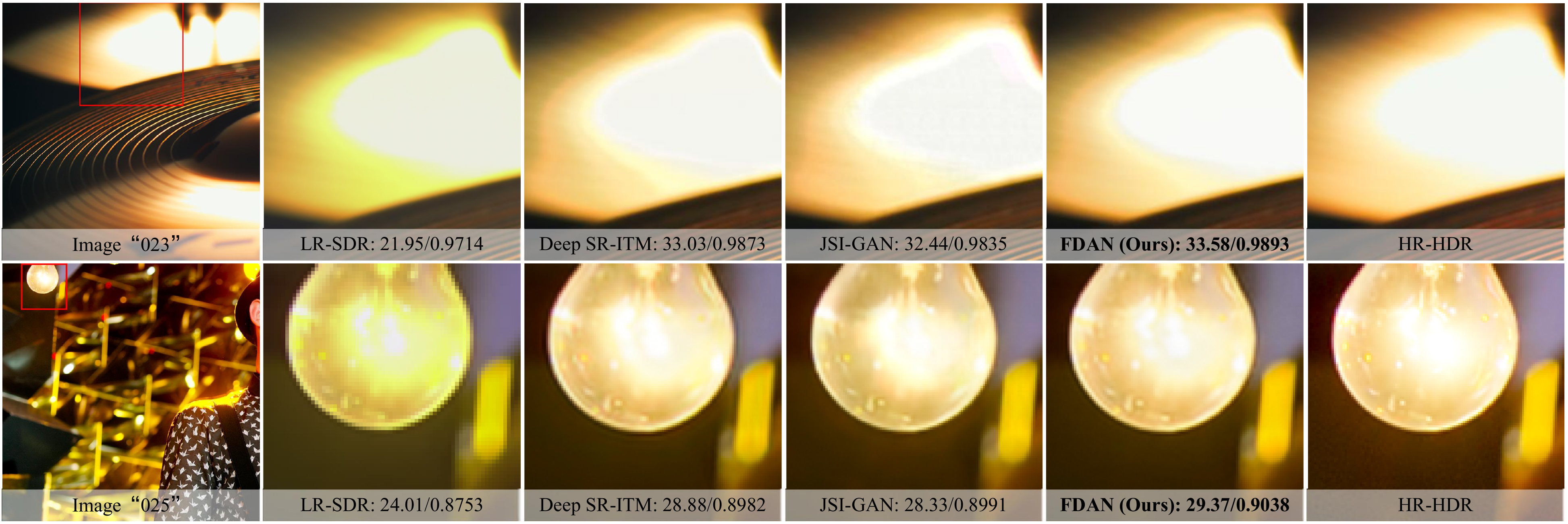}
\caption{\textbf{Visual quality and PSNR (dB)/SSIM results} by different joint SR-ITM methods on dataset in~\cite{kim2019deep} at an SR scale factor of 4.}
\label{fig:experiment_dsi_visualization_scale_04}
\end{figure*}

\noindent
\textbf{Visual quality}.
The comparison results of visual quality with the scale factor 4 are shown in~\figref{fig:experiment_dsi_visualization_scale_04}.
We observe that our FDAN network, designed in a very lightweight structure, obtains comparable or even better visual results when compared to the two competing SRITM methods, \ie, Deep SR-ITM~\cite{kim2019deep} and JSI-GAN~\cite{kim2020jsi}.



\subsection{Ablation Study}
Here, we provide comprehensive experiments to access:
1) effectiveness of our feature decomposition strategy;
%
%
2) the number of FDB blocks in an HFDG group;
3) the number of HFDG groups in our FDAN;
4) how feature aggregation contributes to our FDAN on joint SR-ITM.
All the variants of our FDAN are trained on the SRITM-4K \texttt{training} set and evaluated on the SRITM-4K \texttt{test} set when the scale factor $s=4$.

\noindent
\textbf{1.\ Effectiveness of our feature decomposition strategy}.
To study this problem, we compare our FDAN network with two variants of FDAN on our SRITM-4K dataset at the scale factor of 4:
1) FDAN without feature decomposition: we replace HFDG in our FDAN with the residual block to obtain the variant ``FDAN w/o FD'' which has a similar amount of network parameters as FDAN;
2) FDAN without equally split operation: we remove the split operation in FDB to build the variant, ``FDAN w/o Split'', where the initial base feature $F_{ini}^{base}$ is directly extracted from the the input feature $F_{in}$.
Note that the decomposition cannot be performed with the unequally channel spilt operation due to the mismatch between channel number of $F_{ini}^{base}$ and $F_{in}$.
As shown in~\tableref{table:rebuttal_0}, FDAN outperforms the other competitors on PSNR and SSIM, but with less FLOPs.
This shows that our feature decomposition strategy boosts FDAN on joint SR-ITM on PSNR and SSIM.

\begin{table}[!h]
   \vspace{-0.2cm}
   \begin{center}
     \setlength\tabcolsep{5pt}
     \caption{\textbf{Comparison of methods on the number of parameters (Params), FLOPs, PSNR, and SSIM} on our SRITM-4K dataset.
     In the variant ``FDAN w/o FD'', we replace HFDG in our FDAN with the residual block. 
     The split operation is removed in FDB in the variant ``FDAN w/o Split''.
     }
     \label{table:rebuttal_0}
   \scalebox{0.7}{
     \begin{tabular}{c|c|c|c|c}
        \hline 
        Method & Params(K)$\downarrow$ & FLOPs(G)$\downarrow$ & PSNR(dB)$\uparrow$ & SSIM$\uparrow$ \\
        \hline
        FDAN w/o FD & \textbf{130.70} & 135.24 & 31.05 & 0.9672 \\
        FDAN w/o Split & 568.41 & 557.57 & 31.29 & 0.9673 \\
        \textbf{FDAN (Ours)} & 142.24 & \textbf{117.22} & \textbf{31.75} & \textbf{0.9693} \\
        \hline
     \end{tabular}
  }
  \end{center}
  \vspace{-0.8cm}
\end{table}

\noindent
\textbf{2. How to decide the number of FDB blocks in one HFDG group?}
In each HFDG group, the input feature is decomposed into multi-level fine-grained detail and base features by a total of $B$ FDB blocks.
To find the best $B$, we perform experiments with four variants of our FDAN with $B=1,2,3,4$, respectively.
The results are listed in~\tableref{table:exp_ablation_3}.
Our FDAN achieves the best PSRN and SSIM results when $B=3$.
This demonstrates that our FDAN performs best on joint SR-ITM when decomposing each input feature of our FDB block into 3 levels of detail and base features.\\

\begin{table}[!h]
    \vspace{-0.8cm}
	\begin{center}
		\setlength\tabcolsep{20pt}
		\caption{\textbf{Results of PSNR, SSIM, and FLOPs by our FDAN with different numbers of FDB blocks ($B$) in an HFDG group} on the SRITM-4K \texttt{test} set.}
    \scalebox{0.7}{
		\begin{tabular}{@{\hskip 6pt}c@{\hskip 6pt}|c|c|c}
			\hline 
			$B$ & PSNR (dB)$\uparrow$ & SSIM$\uparrow$ & FLOPs (G)$\downarrow$ \\
			\hline
			1 & 31.39 & 0.9681 & 106.02 \\
			2 & 31.49 & 0.9685 & 114.98 \\
			3 & \textbf{31.75} & \textbf{0.9693} & 117.22 \\
			4 & 31.60 & 0.9685 & 117.78 \\
			\hline
		\end{tabular}
    }
	    \label{table:exp_ablation_3}
    \end{center}
    \vspace{-0.8cm}
\end{table}

\noindent
\textbf{3. How to determine the number of HFDG groups in our FDAN?}
Our FDAN is based on several cascaded HFDG groups.
This naturally raises a question: what is the best number of HFDG groups for our FDAN network?
To answer this question, we perform experiments by our FDAN network with different values of $G=$2, 4, 6, or 8.
The results listed in~\tableref{table:exp_ablation_4} show that our FDAN achieves the best PSNR and SSIM results when $G=6$.
This indicates that cascading $G=6$ HFDG groups in our FDAN network can efficiently exploit multi-scale information for joint SR-ITM.\\

\begin{table}[!h]
	\vspace{-0.8cm}
	\begin{center}
		\setlength\tabcolsep{20pt}
		\caption{\textbf{Results of PSNR, SSIM, and FLOPs by our FDAN with different numbers of HFDGs ($G$)} on the SRITM-4K \texttt{test} set.}
    \scalebox{0.7}{
		\begin{tabular}{@{\hskip 6pt}c@{\hskip 6pt}|c|c|@{\hskip 0pt}r@{\hskip 30pt}}
			\hline 
			$G$ & PSNR (dB)$\uparrow$ & SSIM$\uparrow$ & \multicolumn{1}{c}{FLOPs (G)$\downarrow$} \\
			\hline
			2 & 31.60 & 0.9689 & 68.72 \\
			4 & 31.67 & 0.9685  & 92.97 \\
			6 & \textbf{31.75} & \textbf{0.9693} & 117.22  \\
			8 & 31.71 & 0.9692 & 141.47 \\
			\hline
		\end{tabular}
    }
	    \label{table:exp_ablation_4}
	    \vspace{-0.8cm}
	\end{center}
\end{table}

\noindent
\textbf{4. Does the aggregation of multi-scale features from HFDG groups contributes to our FDAN on joint SR-ITM?}
To answer this question, we remove the feature aggregation and only utilize the feature of the last HFDG group in our FDAN network, resulting in a variant ``FDAN-A''.
%
The comparison results are shown in~\tableref{table:exp_ablation_5}.
One can see that our FDAN outperforms the ``FDAN-A'' by 0.32dB and 0.0006 in terms of PSNR and SSIM.
This demonstrates that the aggregation of HFDG features boosts our FDAN network on joint SR-ITM by exploiting multi-scale information.

\begin{table}[!h]
\begin{center}
\vspace{-0.2cm}
\setlength\tabcolsep{14pt}
\caption{\textbf{Results of PSNR, SSIM, and FLOPs by our FDAN \textsl{w/} or \textsl{w/o} feature aggregation} on the SRITM-4K \texttt{test} set.
``FDAN-A'' denotes the variant, where we remove the feature aggregation and only utilize the feature of the last HFDG group.
}
\scalebox{0.7}{
\begin{tabular}{l|c|c|c}
\hline 
\multirow{1}{*}{\begin{tabular}[c]{@{\hskip 2pt}c@{\hskip 2pt}}Variant\end{tabular}} & PSNR (dB)$\uparrow$ & SSIM$\uparrow$ & FLOPs (G)$\downarrow$ \\
\hline
FDAN-A & 31.43 & 0.9687 & 105.27 \\
FDAN & \textbf{31.75} & \textbf{0.9693}  & 117.22 \\
\hline
\end{tabular}
}
\label{table:exp_ablation_5}
\end{center}
\vspace{-0.8cm}
\end{table}

\section{Conclusion}
\label{sec:Conclusion}
In this paper, we developed a lightweight Feature Decomposition Aggregation Network (FDAN) for efficient joint SR-ITM, generalizing the rigid decomposition by guided filtering techniques on the image domain to a stronger and broader decomposition mechanism on the feature domain.
To well handle fine-grained image details and context, we designed a Feature Decomposition Block (FDB), which achieves learnable separation of feature details and contrasts.
We further cascaded multi-level FDB blocks to build up a Hierarchical Feature Decomposition Group for powerful feature learning capability.
We also constructed a large-scale dataset, \ie, SRITM-4K, to bridge the gap between the scale of dataset in~\cite{kim2019deep} and the power of deep CNNs for joint SR-ITM.
Experimental results on the two datasets demonstrated that, our FDAN outperforms previous state-of-the-art methods on joint SR-ITM, with a significant reduction on model complexity (\eg, number of parameters) and computational costs (\eg, amounts of FLOPs and MACs).









\bibliographystyle{elsarticle-num}
\bibliography{main}

\subsection*{  }
\noindent \textbf{Gang Xu}
received his B.Sc. degree in information security from Xidian University, China in 2018, and Ph.D. degree in computer science and technology from Nankai University, China in 2023.\par

\noindent \textbf{Yu-Chen Yang}
received the B.Sc. degree from the School of Statistics and Data Science, Nankai University, Tianjin, China, in 2022, where he is currently
pursuing the M.Sc. degree with the School of Statistics and Data Science.\par

\noindent \textbf{Liang Wang}
received his Ph.D. degree from the Institute of Automation, Chinese Academy of Sciences (CASIA) in 2004. From 2004 to 2010, he has been working at Imperial College London, United Kingdom, Monash University, Australia, the University of Melbourne, Australia, and the University of Bath, United Kingdom, respectively. Currently, he is a full Professor at the National Lab of Pattern Recognition, CASIA.\par

\noindent \textbf{Xian-Tong Zhen}
received the B.S. and M.E. degrees from Lanzhou University, Lanzhou, China in 2007 and 2010, respectively, and the Ph.D. degree from the Department of Electronic and Electrical Engineering, The University of Sheffield, U.K., in 2013. He is currently with University of Amsterdam and Inception Institute of Artificial Intelligence (IIAI).\par

\noindent \textbf{Jun Xu}
received his B.Sc. and M.Sc. degrees from School of Mathematics Science, Nankai University, Tianjin, China, in 2011 and 2014, respectively, and Ph.D. degree from Department of Computing, Hong Kong Polytechnic University, in 2018.
He worked as a Research Scientist at IIAI, Abu Dhabi, UAE. He is currently an Associate Professor with School of Statistics and Data Science, Nankai University.\par

\end{document}